\documentclass[10pt,twocolumn,letterpaper]{article}

\usepackage{iccv}
\usepackage{times}
\usepackage{epsfig}
\usepackage{graphicx}
\usepackage{amsmath}
\usepackage{amssymb}

\usepackage[font=small,skip=4pt]{caption}
\usepackage[font=small]{subfig}
\usepackage{comment}
\usepackage{multirow}
\usepackage{booktabs}
\usepackage{xcolor}
\usepackage{multirow}
\usepackage{tabularx}
\usepackage[super]{nth}
\newcolumntype{Y}{>{\centering\arraybackslash}X}
\usepackage{enumitem}
\usepackage{algorithm}
\usepackage{algorithmic}
\usepackage{color,soul}

\usepackage{color,soul}
\soulregister\cite7
\soulregister\ref7
\soulregister\pageref7

\usepackage[pagebackref=true,breaklinks=true,letterpaper=true,colorlinks,bookmarks=false]{hyperref}
\usepackage[accsupp]{axessibility}  % Improves PDF readability for those with disabilities.
\iccvfinalcopy % *** Uncomment this line for the final submission

\def\iccvPaperID{3793} % *** Enter the ICCV Paper ID here

\ificcvfinal\fi

\begin{document}
	
%%%%%%%%% TITLE
\title{DC-ShadowNet: Single-Image Hard and Soft Shadow Removal Using  \\
	Unsupervised Domain-Classifier Guided Network}

\author{Yeying Jin$^{1}$, Aashish Sharma$^{1}$, and Robby T. Tan$^{1,2}$\\
	$^1$National University of Singapore, $^2$Yale-NUS College\\
	{\tt\small 
		jinyeying@u.nus.edu, aashish.sharma@u.nus.edu, robby.tan@\{nus,yale-nus\}.edu.sg}
	}
\maketitle
\thispagestyle{empty}
\def\thefootnote{$\dagger$}\footnotetext{This work is supported by   MOE2019-T2-1-130.}\def\thefootnote{\arabic{footnote}}

\maketitle
\ificcvfinal\thispagestyle{empty}\fi

%%%%%%%%% ABSTRACT
\begin{abstract}
Shadow removal from a single image is generally still an open problem. Most existing learning-based methods use supervised learning and require a large number of paired images (shadow and corresponding non-shadow images) for training. A recent unsupervised method, Mask-ShadowGAN~\cite{Hu19}, addresses this limitation. However, it requires a binary mask to represent shadow regions, making it inapplicable to soft shadows. To address the problem, in this paper, we propose an unsupervised domain-classifier guided shadow removal network, DC-ShadowNet. Specifically, we propose to integrate a shadow/shadow-free domain classifier into a generator and its discriminator, enabling them to focus on shadow regions. To train our network, we introduce novel losses based on physics-based shadow-free chromaticity, shadow-robust perceptual features, and boundary smoothness. Moreover, we show that our unsupervised network can be used for test-time training that further improves the results. Our experiments show that all these novel components allow our method to handle soft shadows, and also to perform better on hard shadows both quantitatively and qualitatively than the existing state-of-the-art shadow removal methods.
Our code is available at: \url{https://github.com/jinyeying/DC-ShadowNet-Hard-and-Soft-Shadow-Removal}.
\end{abstract}

%%%%%%%%% BODY TEXT
\section{Introduction}
\label{sec:intr}
\begin{figure}[t]
	\centering
	%\captionsetup[subfigure]{labelformat=empty}
	\captionsetup[subfloat]{farskip=2pt}
	\subfloat[Input Image]
	{\includegraphics[width = 0.49\columnwidth]{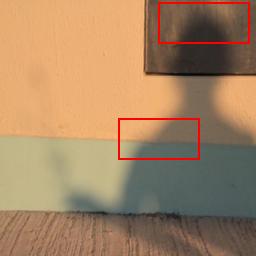}}\hfill
	\subfloat[Ground Truth]
	{\includegraphics[width = 0.49\columnwidth]{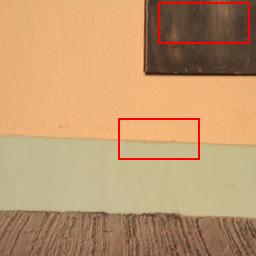}}\hfill\\
	\subfloat[Mask-ShadowGAN~\cite{Hu19}]
	{\includegraphics[width = 0.49\columnwidth]{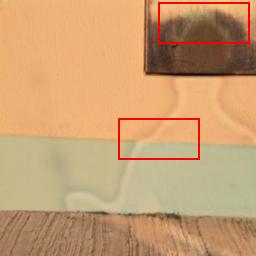}}\hfill
	\subfloat[\textbf{Our DC-ShadowNet}]
	{\includegraphics[width = 0.49\columnwidth]{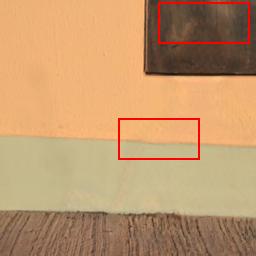}}\hfill
	\caption{Existing state-of-the-art shadow removal methods like Mask-ShadowGAN~\cite{Hu19} fail to remove soft shadows properly and create artifacts (see regions inside red boxes). Compared to it, our method generates a better shadow-free output.}
	\vspace{-0.15in}
	\label{fig:intro}
\end{figure}

Shadow removal from a single image can benefit many applications, such as image editing, scene relighting, etc., \cite{Lalonde09,Khan15,Kawakami05}. 
Unfortunately, in general, removing shadows from a single image is still an open problem.
Existing physics-based methods for shadow removal~\cite{Finlayson05,Finlayson09,Guo11} are based on entropy minimization that can capture the invariant features of shadow and non-shadow regions belong to the same surfaces in the log-chromaticity space.  
These methods, however, tend to fail, particularly when the image surfaces are close to achromatic (e.g. gray or white surfaces), and are not designed to handle soft shadow images.

Unlike physics-based methods, deep-learning methods, e.g.  \cite{Qu17,Wang18,Hu18,Le19,Cun20,le2020shadow}, are more robust to different conditions of image surfaces and lighting.
However, most of these methods are based on fully-supervised learning, which means that for training, they require pairs of shadow and their corresponding non-shadow images. To collect these image pairs in a large amount, particularly for images containing diverse scenes and shadows can be considerably expensive.

Recently, Hu~\etal propose an unsupervised method, Mask-ShadowGAN~\cite{Hu19}, the network architecture of which is based on CycleGAN~\cite{Zhu17}. To remove shadows, the method mainly relies on adversarial training that employs a discriminator to check the quality of the generated output. Unfortunately, due to the absence of ground truth, the discriminator relies solely on unpaired non-shadow images, which can cause the generator to produce incorrect outputs. 
Moreover, the method uses a binary mask to represent shadow regions present in the input image, making it inapplicable to soft shadow images. Fig.~\ref{fig:intro} shows an example where for the given soft-shadow input image, the output generated by the method~\cite{Hu19} is improper.

In this paper, our goal is to remove both hard and soft shadows from a single image. 
To achieve this, we propose {\it DC-ShadowNet}, an unsupervised network guided by the shadow/shadow-free domain classifier. 
Specifically, we integrate a domain classifier (that classifies the input image to either shadow or shadow-free domain) into our generator and its corresponding discriminator.
This allows our generator and discriminator to focus on shadow regions and thus perform better shadow removal.
Unlike the existing unsupervised method~\cite{Hu19}, which only relies on adversarial training based on an unpaired discriminator (i.e. using unpaired non-shadow images as reference images), our method uses additional novel unsupervised losses that enable our method to achieve better shadow removal results. 
Our new losses are based on physics-based shadow-free chromaticity, shadow-robust perceptual features, and boundary smoothness.

Our physics-based shadow-free chromaticity loss employs a shadow-free chromaticity image, which is obtained from the input shadow image by performing entropy minimization in the log-chromaticity space~\cite{Finlayson05}.
Our shadow-robust perceptual features loss uses shadow-robust features obtained from the input shadow image using the pre-trained VGG-16 network~\cite{Johnson16}. 
We also add a boundary smoothness loss to ensure that our output shadow-free image has smoother transitions in the regions that contained shadow boundaries. 
All these ideas enable our method to better deal with hard and soft shadow images compared to existing methods like~\cite{Hu19} (see Fig.~\ref{fig:intro} for an example showing the better performance of our method).   
Furthermore, we show that our method being unsupervised can be used for test-time training to further improve the performance of our method. 
As a summary, here are our contributions:
\begin{enumerate}[noitemsep,topsep=1pt]
	\item We introduce DC-ShadowNet, a new unsupervised single-image shadow removal network guided by a domain classifier  to focus on shadow regions.  
 	\item We propose novel unsupervised losses based on physics-based shadow-free chromaticity, shadow-robust perceptual features, and boundary smoothness losses for robust shadow removal.
 	\item To our knowledge, our method is the first unsupervised method to perform shadow removal robustly for both hard and soft shadow in a single image.
\end{enumerate}
 
%------------------------------------------------------------------------
\section{Related work}
\label{sec:related}
Physics-based shadow removal methods (e.g.~\cite{Finlayson01,Drew03,Finlayson04,Finlayson05,Finlayson09}) are based on the physics models of illumination and surface colors. 
These methods assume that the surface colors in the input image are chromatic, and hence they are erroneous when this assumption does not hold. 
These methods are designed to remove hard shadows only. 
In contrast, our method is based on unsupervised learning and is designed to handle both hard and soft shadows. 
Also, our method is more robust in dealing with achromatic surfaces.

Some other non-learning-based methods rely on user interaction. Gryka \etal \cite{Gryka15} propose a regression model to learn a mapping function of shadow image regions and their corresponding shadow mattes. However, they need the user to provide brush strokes to relight shadow regions.  
Guo \etal \cite{Guo11,Guo12} use annotated ground truth to learn the appearances of shadow regions. 
Unlike these methods, our method is learning-based and does not rely on hand-crafted feature descriptors, making it more robust. Moreover, our method does not need any annotated ground truth and user interaction; hence, it is more practical and efficient.

To address the aforementioned limitations of non-deep learning methods, many deep learning methods are proposed. Wang \etal~\cite{Wang18} use a stacked conditional GAN (ST-CGAN) to detect and remove shadows jointly. 
Le \etal~\cite{Le19,le2020shadow} propose SP+M-Net do shadow removal using image decomposition.
Hu \etal~\cite{Hu18,hu2019direction} propose to add global and direction-aware context into the direction-aware spatial context (DSC) module.
Ding \etal~\cite{Ding19} introduce an LSTM-based attentive recurrent GAN (ARGAN) to detect and remove shadows. 
All these methods are trained on paired data using supervised learning. 
Hence, training them using various soft shadows and complex scenes is difficult, since obtaining the ground truths is intractable.
In contrast, our method is based on unsupervised learning and does not need any paired data.

\begin{figure*}
	\vspace{-0.1in}
	\begin{center}
		\captionsetup[subfigure]{labelformat=empty}
		{\includegraphics[width=17.7cm,height=8cm]{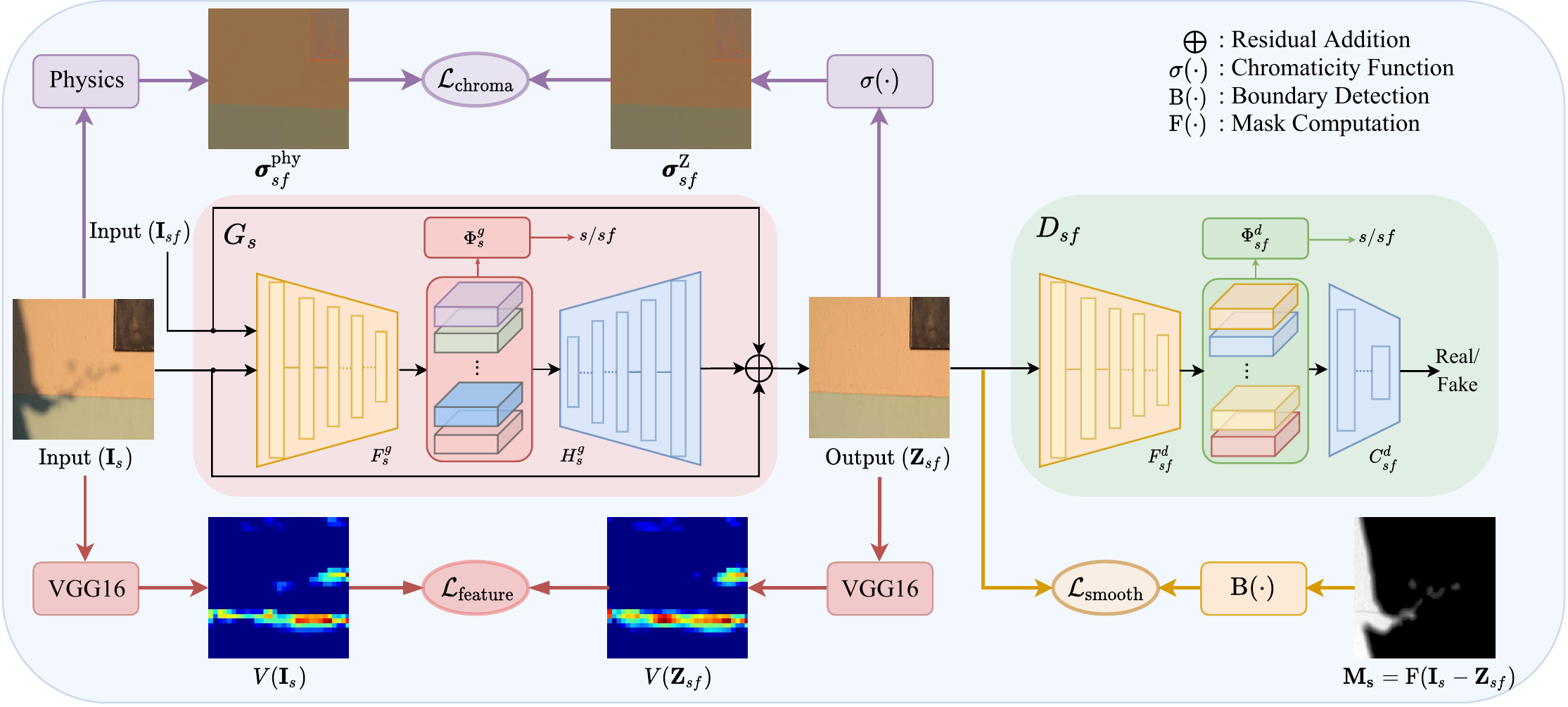}}
		\vspace{-0.125in}
	\end{center}
	\vspace{-0.15in}
	\caption{\textbf{Network Architecture of Our DC-ShadowNet.} 
		We have two domains: shadow, $s$, and shadow-free, $sf$. Our shadow removal generator is represented by $G_s$. It consists of an encoder $F^g_s$, a decoder $H^g_s$, and a domain classifier $\Phi^g_s$. 
		We also use a discriminator $D_{sf}$ that consists of its own encoder $F^d_{sf}$, a classifier $C^d_{sf}$ and a domain classifier $\Phi^d_{sf}$.
		For the input shadow image $\mathbf{I}_s$, its corresponding output shadow-free image is represented by $\mathbf{Z}_{sf}$. 
		Also, for the unpaired input shadow-free image $\mathbf{I}_{sf}$, $G_{s}$ reconstruct the image back.
		The domain classifiers, $\Phi^g_s$ and $\Phi^d_{sf}$, are used to classify whether the inputs to their respective networks, $G_s$ and $D_{sf}$, belong to shadow ($s$) or shadow-free ($sf$) domain.
		To guide our generator $G_s$ to do shadow removal, other than adversarial loss from the discriminator $D_{sf}$, we include novel losses: shadow-free chromaticity loss $\mathcal{L}_{\text{chroma}}$ (purple) guided by the physics-based shadow-free chromaticity $\pmb{\sigma}_{sf}^\text{phy}$ obtained from $\mathbf{I}_s$; shadow-robust feature loss $\mathcal{L}_{\text{feature}}$ (red) guided by the shadow-robust perceptual features $V(\mathbf{I}_s)$ obtained from $\mathbf{I}_s$, and boundary smoothness loss $\mathcal{L}_{\text{smooth}}$ (orange) guided by the boundary detection of our generated soft shadow mask $\mathbf{M}_s$.}
	\vspace{-0.1in}
	\label{fig:network}
\end{figure*}

Recently, Hu \etal~\cite{Hu19} propose an unsupervised deep-learning method Mask-ShadowGAN. Unfortunately, since it mainly relies on adversarial training for shadow removal, it cannot guarantee that the generated output images are shadow-free since there is no strong guidance for the network to do so. Moreover, it cannot handle soft shadows due to the use of binary masks. 
In contrast, our method DC-ShadowNet uses new additional unsupervised losses and domain-classifier guided network that helps our method to more effectively deal with hard and soft shadows.
%

%-------------------------------------------------------------------------
\section{Proposed Method}
\label{sec:method}
%-------------------------------------------------------------------------
Fig.~\ref{fig:network} shows the architecture of our network, DC-ShadowNet.  
Given a shadow input image, $\mathbf{I}_s$, we use a generator, $G_s$, to transform it into a shadow-free output image $\mathbf{Z}_{sf}$. 
Also, given an unpaired shadow-free input image, $\mathbf{I}_{sf}$, we expect the generator, $G_{s}$, to simply  reconstruct the image back.
Therefore, the generator $G_s$, whether its input is a shadow or shadow-free image,  always generates a shadow-free output image. 
Note that, in our method, we have two domains: shadow, $s$, and shadow-free, $sf$. 

Our generator $G_s$ consists of an encoder ($F^g_s$), decoder ($H^g_s$) and a domain classifier ($\Phi^g_s$). 
We use a discriminator $D_{sf}$ to assess the quality of the shadow removal output. It consists of an encoder ($F^d_{sf}$), a classifier ($C^d_{sf}$) and a domain classifier ($\Phi^d_{sf}$).
Both the domain classifiers, $\Phi^g_s$ and $\Phi^d_{sf}$, are used to classify the inputs of their respective modules, $G_s$ and $D_{sf}$, belonging to either shadow or shadow-free domain. However, unlike $\Phi^g_s$, which is trained together with $G_s$, $\Phi^d_{sf}$ is pre-trained, and its weights are kept frozen while training $D_{sf}$. 
The underlying idea of integrating the domain classifier into our generator and its discriminator is to guide our network to focus on shadow regions.
The reference images of our discriminator are the unpaired shadow-free real images. Our discriminator's classifier, $C^d_{sf}$, outputs the real/fake binary label, where real refers to the label given to an image that belongs to the reference images. 

While not shown in Fig.~\ref{fig:network}, for the sake of clarity, we employ another generator $G_{sf}$ and the shadow mask to transform the shadow-free output image back to a shadow image, in order to enforce reconstruction consistency~\cite{Zhu17} and locate the shadow regions.
Also, another discriminator $D_{s}$ is used to distinguish whether the generated shadow image is real or not.
Our method, DC-ShadowNet, is trained in an unsupervised manner using our losses, which are described in the following sections.
%

%-------------------------------------------------------------------------
\subsection{Shadow-Free Chromaticity Loss}
Given a shadow input image $\mathbf{I}_s$, we obtain a physics-based shadow-free chromaticity image $\pmb{\sigma}_{sf}^\text{phy}$, which is used to guide our shadow removal generator $G_s$, through our shadow-free chromaticity loss function. 
Obtaining $\pmb{\sigma}_{sf}^\text{phy}$ from $\mathbf{I}_s$ requires two steps: (1) Entropy Minimization, and (2) Illumination Compensation. 

\begin{figure}[t]
	\begin{center}
		\captionsetup[subfigure]{labelformat=empty}
		{\includegraphics[width=0.49\textwidth]{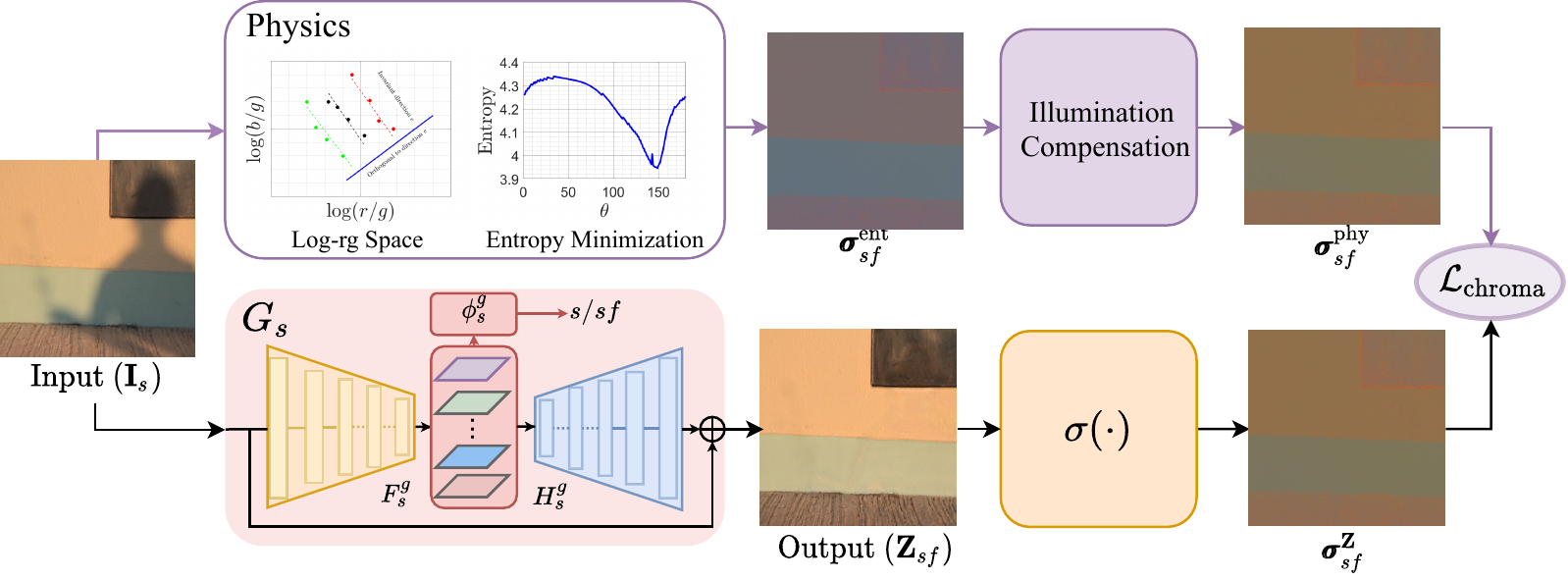}}
		\vspace{-0.125in}	
	\end{center}
	\vspace{-0.2in}
	\caption{\textbf{Shadow-Free Chromaticity Loss}. The upper part is the physics-based pipeline where we use entropy minimization followed by illumination compensation to generate the shadow-free chromaticity image $\pmb{\sigma}_{sf}^\text{phy}$ from the input image $\mathbf{I}_s$.
	The lower part shows our shadow removal generator $G_s$ guided by $\pmb{\sigma}_{sf}^\text{phy}$ through our shadow-free chromaticity loss $\mathcal{L}_\text{chroma}$.}
	\label{fig:chromaticityloss}
	\vspace{0.05in}
\end{figure}

\begin{figure}[t]
	\centering
	%\captionsetup[subfigure]{font=small, labelformat=empty}
	\captionsetup[subfloat]{farskip=2pt}
	\subfloat{\includegraphics[width = 0.196\columnwidth]{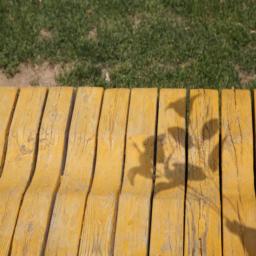}}\hfill
	\subfloat{\includegraphics[width = 0.196\columnwidth]{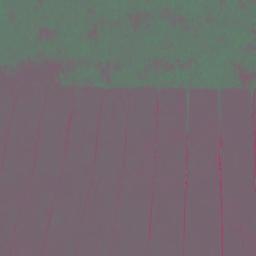}}\hfill
	\subfloat{\includegraphics[width = 0.196\columnwidth]{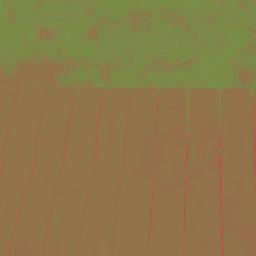}}\hfill
	\subfloat{\includegraphics[width = 0.196\columnwidth]{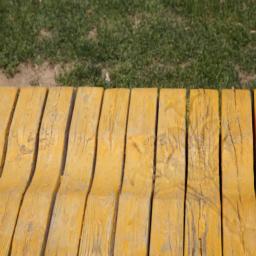}}\hfill
	\subfloat{\includegraphics[width = 0.196\columnwidth]{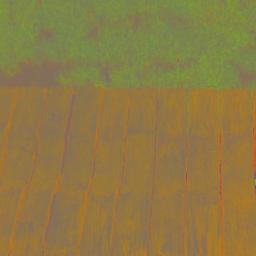}}\hfill\\
	\vspace{-0.015in}
	\setcounter{subfigure}{0}
	\subfloat[Input\label{fig:chromaticity_inp}]	
	{\includegraphics[width = 0.196\columnwidth]{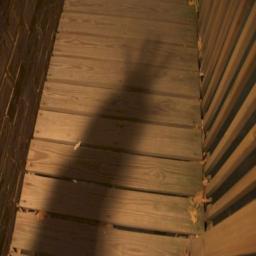}}\hfill
	\subfloat[$\pmb{\sigma}_{sf}^{\text{ent}}$\label{fig:chromaticity_sfchrom1}]	
	{\includegraphics[width = 0.196\columnwidth]{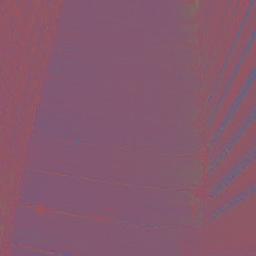}}\hfill
	\subfloat[$\pmb{\sigma}_{sf}^\text{phy}$\label{fig:chromaticity_sfchrom2}]
	{\includegraphics[width = 0.196\columnwidth]{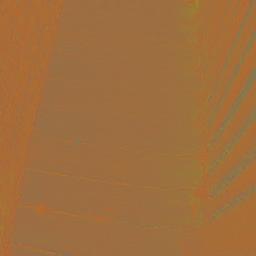}}\hfill
	\subfloat[Output\label{fig:chromaticity_out}]
	{\includegraphics[width = 0.196\columnwidth]{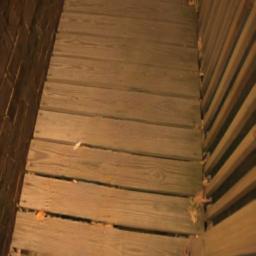}}\hfill
	\subfloat[$\pmb{\sigma}_{SF}^\text{Z}$\label{fig:chromaticity_outchrom}]
	{\includegraphics[width = 0.196\columnwidth]{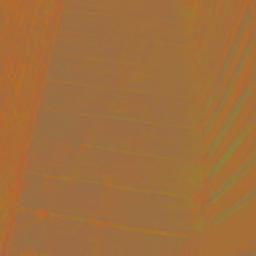}}\hfill\\
	\vspace{-0.01in}
	\caption{\textbf (a) Input shadow image ${\mathbf{I}_{s}}$,  (b) Shadow-free chromaticity after entropy minimization $\pmb{\sigma}_{sf}^{\text{ent}}$, (c) Shadow-free chromaticity after illumination compensation $\pmb{\sigma}_{sf}^{\text{phy}}$, (d) Output shadow-free image $\mathbf{Z}_{sf}$, and (e) Chromaticity map the of output image $\pmb{\sigma}_{sf}^\textbf{Z}$. Our shadow-free chromaticity loss constrains (e) to be similar to (c) facilitating better shadow removal.}
	\vspace{-0.1in}
	\label{fig:chromaticity}
\end{figure}

\vspace{0.2cm}
\noindent \textbf{Entropy Minimization} 
Following~\cite{Finlayson09}, as shown in Fig.~\ref{fig:chromaticityloss}, we plot the input shadow image $\mathbf{I}_s$ onto the log-chromaticity space, calculate the entropy, and use the entropy minimization to find the projection direction $\theta$, which is specific to $\mathbf{I}_s$. 
From $\theta$, we can obtain a shadow-free chromaticity map $\pmb{\sigma}_{sf}^{\text{ent}}$ that no longer contains any shadows (see Figs.~\ref{fig:chromaticityloss} and \ref{fig:chromaticity_sfchrom1}).
However, owing to the projection, there is a color shift present in $\pmb{\sigma}_{sf}^{\text{ent}}$, which can be corrected by using the illumination compensation procedure.

\vspace{0.2cm}
\noindent \textbf{Illumination Compensation} 
To correct the color of the shadow-free chromaticity map $\pmb{\sigma}_{sf}^{\text{ent}}$, following~\cite{Drew03}, we add back the original illumination color of the non-shadow regions to the map.
For this, we use uniformly sampled $30\%$ of the brightest pixels from the input image $\mathbf{I}_s$ based on the assumption that these pixels are located in the non-shadow regions of $\mathbf{I}_s$. 
Once we reinstate the illumination color, we can obtain a new shadow-free chromaticity map $\pmb{\sigma}_{sf}^\text{phy}$,  (see Figs.~\ref{fig:chromaticityloss} and \ref{fig:chromaticity_sfchrom2}).

Having obtained our shadow-free chromaticity, $\pmb{\sigma}_{sf}^\text{phy}$, for the output shadow-free image $\mathbf{Z}_{sf}$, we compute its chromaticity map $\pmb{\sigma}_{sf}^\mathbf{Z}$ by:
\begin{equation}
	\pmb{\sigma}_{{sf}_c}^\mathbf{Z}\!= \!\frac{\mathbf{Z}_{{sf}_c}}{(\mathbf{Z}_{{sf}_r} + \mathbf{Z}_{{sf}_g} + \mathbf{Z}_{{sf}_b})},
	\label{eq:rgchromaticity}
\end{equation}
where $c\in\{r,g,b\}$ represents a color channel, $\mathbf{Z}_{sf} = [\mathbf{Z}_{{sf}_r}, \mathbf{Z}_{{sf}_g}, \mathbf{Z}_{{sf}_b}]$, and $\pmb{\sigma}_{sf}^\mathbf{Z} = [\pmb{\sigma}_{{sf}_r}^\mathbf{Z}, \pmb{\sigma}_{{sf}_g}^\mathbf{Z}, \pmb{\sigma}_{{sf}_b}^\mathbf{Z}]$. We can now define our shadow-free chromaticity loss as:
\begin{equation}
	\mathcal{L}_{\rm chroma}(G_s) = \mathbb{E}_{\mathbf{I}_s} 
	\big[||\pmb{\sigma}_{sf}^\mathbf{Z} - \pmb{\sigma}_{sf}^\text{phy}||_{1}\big].
	\label{eq:loss_ch}
\end{equation}
Using the loss function expressed in Eq.~(\ref{eq:loss_ch}), we enforce the chromaticity of the output shadow-free image, $\pmb{\sigma}_{sf}^\mathbf{Z}$, to be the same as our physics-based shadow-free chromaticity $\pmb{\sigma}_{sf}^\text{phy}$, which can be observed in the results shown in Fig.~\ref{fig:chromaticity} for both hard shadow and soft shadow images\footnote{For surfaces that are close to being achromatic, the entropy minimization can fail, which can lead to the improper recovery of the shadow-free chromaticity map. However, due to the presence of our other unsupervised losses, our method can still generate proper shadow removal results.}.

\begin{figure}[t]
	\centering
	%\captionsetup[subfigure]{labelformat=empty}
	\captionsetup[subfloat]{farskip=2pt}
	\subfloat{\includegraphics[width = 0.245\columnwidth]{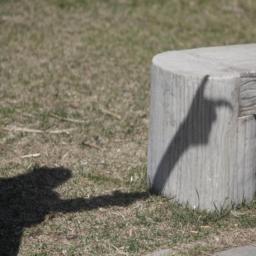}}\hfill
	\subfloat{\includegraphics[width = 0.245\columnwidth]{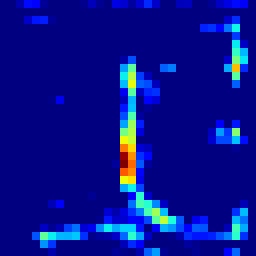}}\hfill
	\subfloat{\includegraphics[width = 0.245\columnwidth]{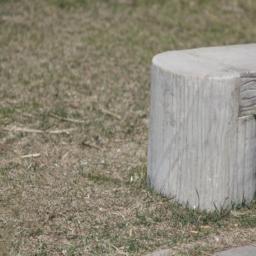}}\hfill
	\subfloat{\includegraphics[width = 0.245\columnwidth]{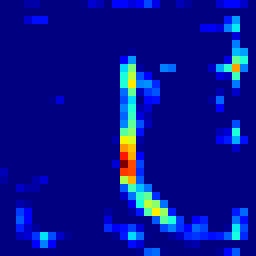}}\hfill\\
	\vspace{-0.01in}
	\setcounter{subfigure}{0}
	\subfloat[Input]
	{\includegraphics[width = 0.245\columnwidth]{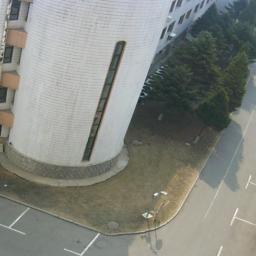}}\hfill
	\subfloat[$V(\mathbf{I}_s)$]
	{\includegraphics[width = 0.245\columnwidth]{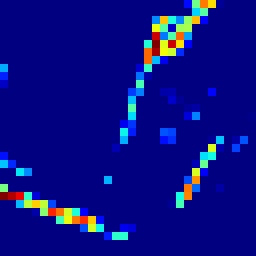}}\hfill
	\subfloat[Output]
	{\includegraphics[width = 0.245\columnwidth]{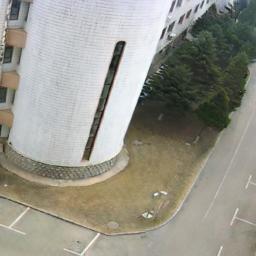}}\hfill
	\subfloat[$V(\mathbf{Z}_{sf})$]
	{\includegraphics[width = 0.245\columnwidth]{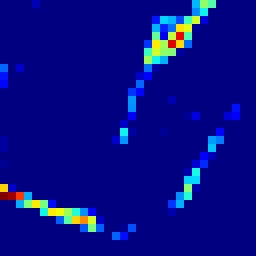}}\hfill\\
	\vspace{-0.01in}
	\caption{(a) Input shadow image $\mathbf{I}_{s}$, (b) Sample feature map for $\mathbf{I}_{s}$, (c) Output shadow-free image $\mathbf{Z}_{sf}$, and (d) Sample feature map for $\mathbf{Z}_{sf}$. We can observe that features in (b) for the input shadow images are less affected by shadows, and they are similar to the features in (d) owing to our shadow-robust feature loss.}
	%\vspace{-0.1in}
	\label{fig:featureloss}
\end{figure}
%-------------------------------------------------------------------------
\subsection{Shadow-Robust Feature Loss}
Our shadow-robust feature loss is based on the perceptual features obtained from the pre-trained VGG-16 network~\cite{Johnson16, sharma2020nighttime}. 
Since we do not have ground truth to obtain the correct shadow-free features, to guide the shadow-free output, we use features from the input shadow image itself. 
Our underlying idea is that, since with some degree of shadows and lighting conditions, object classification using the pre-trained VGG-16 is known to be robust~\cite{webster2018psyphy}, there should be some features in the pre-trained VGG-16 that are less affected by shadows. Based on this, we perform a calibration experiment and find that the Conv22 layer in the VGG-16 network provides features that are least affected by shadows. 

Hence, from the input shadow image, we obtain the shadow-robust features and use them to guide our shadow-free output image. 
Specifically, given an input shadow image $\mathbf{I}_s$ and the corresponding shadow-free output image $\mathbf{Z}_{sf}$, we define our shadow-robust feature loss as:
\begin{equation}
	\mathcal{L}_{\rm feature}(G_s) = \mathbb{E}_{\mathbf{I}_s}
	[\big\lVert V(\mathbf{Z}_{sf}) - V(\mathbf{I}_s)\big\rVert_1],
	\label{eqn_inv_feature}
\end{equation}
where $V(\mathbf{I}_s)$ and $V(\mathbf{Z}_{sf})$ denote the feature maps extracted from the Conv22 layer of the pre-trained VGG-16 network for $\mathbf{I}_s$ and $\mathbf{Z}_{sf}$ respectively. Fig.~\ref{fig:featureloss} shows some examples where we can observe that the features $V(\mathbf{I}_s)$ are less affected by shadows and represent more of structural information (like edges).

%-------------------------------------------------------------------------
\begin{figure}[t]
	\begin{center}
		\captionsetup[subfigure]{labelformat=empty}
		{\includegraphics[width=0.49\textwidth]{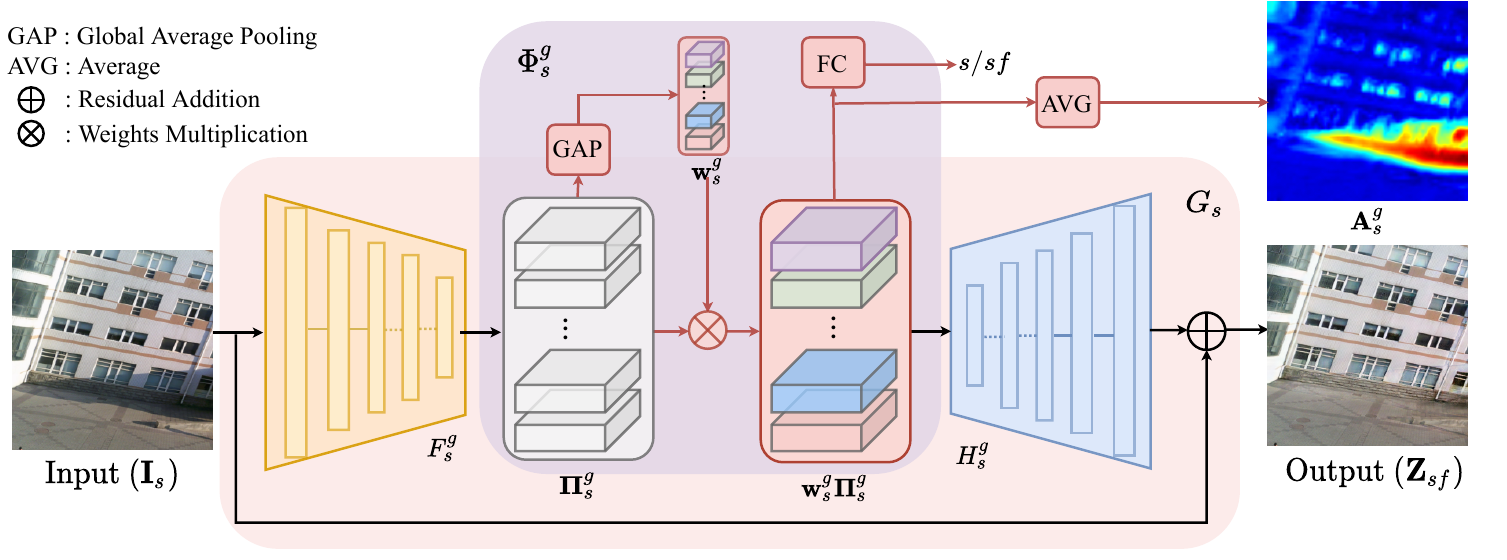}}
		\vspace{-0.125in}	
	\end{center}
	\vspace{-0.2in}
	\caption
	{\textbf{Domain Classification and Shadow Attention.} 
	In the generator $G_s$, its encoder $F^g_s$ extracts feature maps $\mathbf{\Pi}^g_s$ from the input shadow image $\mathbf{I}_s$. As in~\cite{zhou2016learning}, using global average pooling (GAP), the domain classifier $\Phi^g_s$ is trained to learn the weights $\mathbf{w}^g_s$ of the feature maps. Averaging the weighted feature maps generates an attention map $\mathbf{A^g_{s}}$, i.e. $\mathbf{A}^g_s = \frac{1}{n}\sum_{i=1}^{n}{\mathbf{w}^g_s}_i {\mathbf{\Pi}^g_s}_i$ ($n$ being the total number of feature maps), which clearly shows that the network is focusing on shadow regions.}
	\label{fig:camattention}
	\vspace{-0.1in}
\end{figure}

\subsection{Domain Classification Loss}
\begin{figure}[t]
	\centering
	\captionsetup[subfigure]{labelformat=empty}
	\captionsetup[subfloat]{farskip=2pt}
	\rotatebox{90}{\small \phantom{z}\textbf{(a)} Input}\hspace{0.08cm}
	{\includegraphics[width = 0.186\columnwidth]{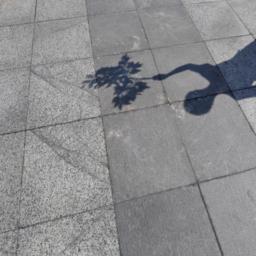}}\hfill
	\subfloat{\includegraphics[width = 0.186\columnwidth]{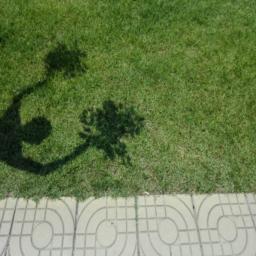}}\hfill
	\subfloat{\includegraphics[width = 0.186\columnwidth]{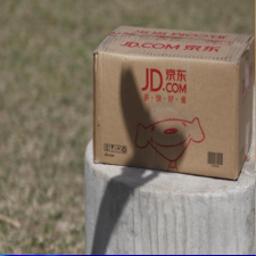}}\hfill
	\subfloat{\includegraphics[width = 0.186\columnwidth]{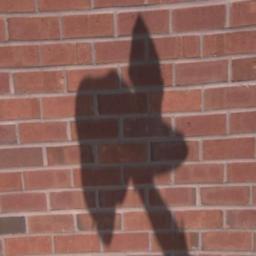}}\hfill
	\subfloat{\includegraphics[width = 0.186\columnwidth]{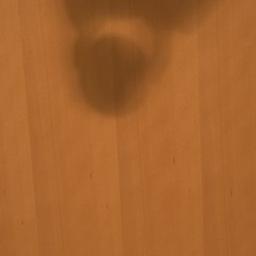}}\hfill\\
	\vspace{-0.015in}	
	\rotatebox{90}{\small \phantom{zz}\textbf{(b)} $\mathbf{A}^g_s$}\hspace{0.08cm}
	{\includegraphics[width = 0.186\columnwidth]{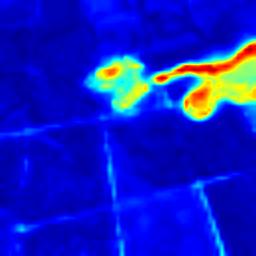}}\hfill
	\subfloat{\includegraphics[width = 0.186\columnwidth]{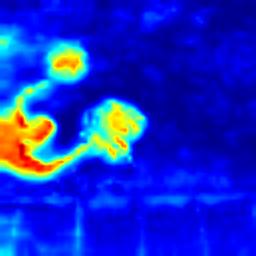}}\hfill
	\subfloat{\includegraphics[width = 0.186\columnwidth]{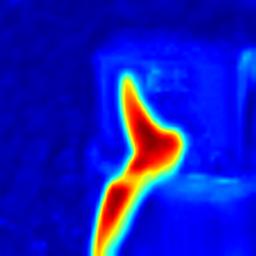}}\hfill
	\subfloat{\includegraphics[width = 0.186\columnwidth]{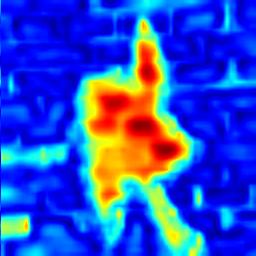}}\hfill
	\subfloat{\includegraphics[width = 0.186\columnwidth]{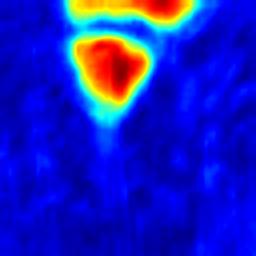}}\hfill\\
	\vspace{-0.015in}
	\rotatebox{90}{\small \phantom{z}\textbf{(c)} Output}\hspace{0.08cm}
	{\includegraphics[width = 0.186\columnwidth]{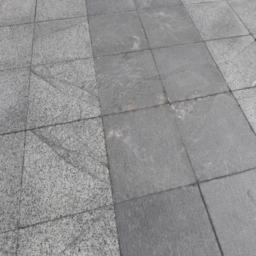}}\hfill
	\subfloat{\includegraphics[width = 0.186\columnwidth]{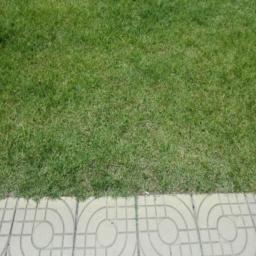}}\hfill
	\subfloat{\includegraphics[width = 0.186\columnwidth]{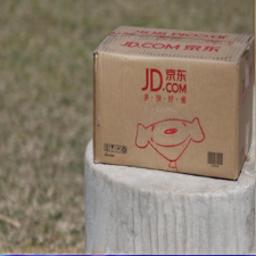}}\hfill
	\subfloat{\includegraphics[width = 0.186\columnwidth]{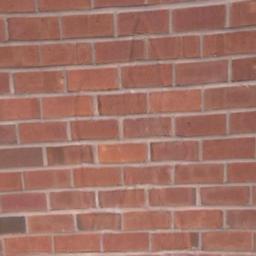}}\hfill
	\subfloat{\includegraphics[width = 0.186\columnwidth]{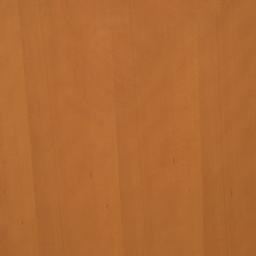}}\hfill\\	
	\caption{ (a) Input shadow image $\mathbf{I}_s$, (b) Attention map $\mathbf{A}^g_s$, and (c) Output shadow-free image $\mathbf{Z}_{sf}$. The attention maps clearly indicate the shadow regions of the input shadow images.}
	%\vspace{-0.1in}
	\label{fig:attentionmaps}
\end{figure}

We incorporate an attention mechanism that allows our DC-ShadowNet to know the shadow removal/restoration regions~\cite{zhou2016learning, qian2018attentive, Kim2020U-GAT-IT:}. 
To achieve this, we create a domain classifier $\Phi^g_s$ and integrate it with the generator $G_s$. We train $\Phi^g_s$ to classify whether the input to $G_s$ is from the shadow or shadow-free domain. Fig.~\ref{fig:camattention} shows the integration of $\Phi^g_s$ into $G_s$ to obtain an attention map $\mathbf{A^g_{s}}$ that highlights shadow regions. 
We also add a similar domain classifier $\Phi^d_{sf}$ to the discriminator $D_{sf}$.
This allows our network to selectively focus on important shadow regions and generate better shadow removal results (see Fig.~\ref{fig:attentionmaps}).

Since the generator can accept either a shadow or shadow-free image as input, it allows us to train it together with its domain classifier. However, for the discriminator, the domain of its input image, which is the output of the generator, can be ambiguous\footnote{In the early stage of training, shadow removal can be improper, and the output of the generator can still have shadows. Hence, it is difficult to ensure that the domain of the output is always shadow-free.}.
For this reason, we pre-train the domain classifier of the discriminator using the following classification loss:
\begin{align}
	\mathcal{L}_{\rm domcls}(D_{sf}) =  
	& \mathbb{E}_{\mathbf{I}_{s}} \Big[-\log\big(\Phi^d_{sf}(F^d_{sf}(\mathbf{I}_{s}))\big)\Big]
	+\nonumber\\& \mathbb{E}_{\mathbf{I}_{sf}} \Big[-\log\big(1-\Phi^d_{sf}(F^d_{sf}(\mathbf{I}_{sf}))\big)\Big],
	\label{eq:loss_cam_d}
\end{align}
and after pre-training, we freeze its weights during the main training cycle that trains our entire network (see Fig.~\ref{fig:network}). To train the domain classifier of the generator, we use a similar classification loss:
\begin{align}
	\mathcal{L}_{\rm domcls}(G_s) =  
	& \mathbb{E}_{\mathbf{I}_s} \Big[-\log\big(\Phi^g_{s}(F^g_s(\mathbf{I}_s))\big)\Big]
	+\nonumber\\& \mathbb{E}_{\mathbf{I}_{sf}} \Big[-\log(1-\Phi^g_{s}(F^g_s(\mathbf{I}_{sf})\big)\Big].
	\label{eq:loss_cam_g}
\end{align}
 
%------------------------------------------------------------------------
\subsection{Boundary Smoothness Loss}
\begin{figure}[t]
	\centering
	%\captionsetup[subfigure]{labelformat=empty}
	\captionsetup[subfloat]{farskip=2pt}
	\subfloat{\includegraphics[width = 0.245\columnwidth,height=1.8cm]{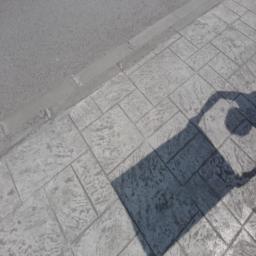}}\hfill
	\subfloat{\includegraphics[width = 0.245\columnwidth,height=1.8cm]{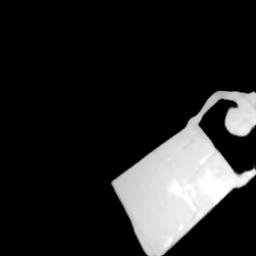}}\hfill
	\subfloat{\includegraphics[width = 0.245\columnwidth,height=1.8cm]{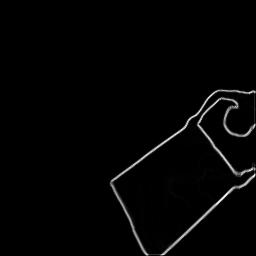}}\hfill
	\subfloat{\includegraphics[width = 0.245\columnwidth,height=1.8cm]{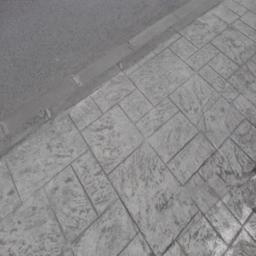}}\hfill\\
	\vspace{-0.01in}
	\setcounter{subfigure}{0}	
	\subfloat[Input\label{fig:smoothloss_input}]
	{\includegraphics[width = 0.245\columnwidth,height=1.8cm]{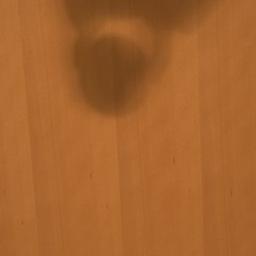}}\hfill
	\subfloat[$\mathbf{M}_s$\label{fig:smoothloss_mask}]
	{\includegraphics[width = 0.245\columnwidth,height=1.8cm]{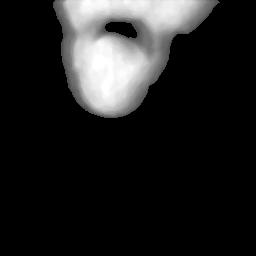}}\hfill
	\subfloat[$\text{B}(\mathbf{M}_s)$\label{fig:smoothloss_bounds}]
	{\includegraphics[width = 0.245\columnwidth,height=1.8cm]{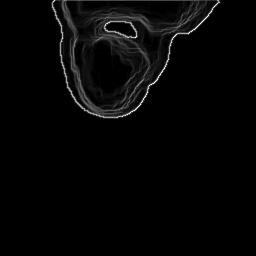}}\hfill
	\subfloat[Output\label{fig:smoothloss_output}]
	{\includegraphics[width = 0.245\columnwidth,height=1.8cm]{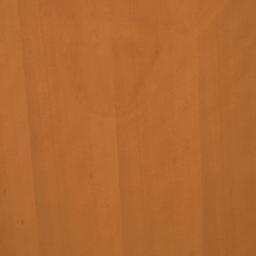}}\hfill\\
	\vspace{-0.01in}
	\caption{(a) Input shadow image, (b) Soft shadow mask $\mathbf{M}_s$, (c) Detected shadow boundaries, and (d) Our output shadow-free results. We can observe that our boundary smoothness loss helps in having smoother outputs in the shadow boundary regions.}
	\vspace{-0.1in}
	\label{fig:smoothloss}
\end{figure}

To ensure that the output shadow-free image $\mathbf{Z}_{sf}$ have smoother transitions in the boundaries defined by the shadow regions of the input shadow image $\mathbf{I}_s$, we also use a boundary smoothness loss:
\begin{equation}
	%\thickmuskip=0mu
	\mathcal{L}_{\rm smooth}(G_s) =
	\mathbb{E}_{\mathbf{I}_s}\Big[\big\lVert \text{B}(\mathbf{M}_s) *
	|\nabla(\mathbf{Z}_{sf})|\rVert_1\Big], 
	\label{eqn_smooth}
\end{equation}
where $\nabla$ is the gradient operation, $\text{B}$ is a noise-robust function~\cite{xu2012structure, sharma2018into, yan2020nighttime} to compute the boundaries of the shadow regions from our shadow mask $\mathbf{M}_s$. 
To obtain $\mathbf{M}_s$, we compute the difference between the input shadow image $\mathbf{I}_s$ and output shadow-free image $\mathbf{Z}_{sf}$, and apply our mask detection function $\text{F}$ on the difference:
\begin{equation}
\mathbf{M}_s = \text{F}\bigl({\mathbf{I}_s}_c - {\mathbf{Z}_{sf}}_c\bigr)\!=\! 
\sum_{c\in\{r,g,b\}}\frac{1}{3}\Big\lvert\bigl(\text{N}({\mathbf{I}_s}_c - {\mathbf{Z}_{sf}}_c)\bigr)\Big\rvert,
\label{eq:mask}
\end{equation}
where the function $\text{N}$ is a normalization function defined as $\text{N}(\mathbf{I}) = (\mathbf{I}- \mathbf{I}_\text{min})/(\mathbf{I}_\text{max} - \mathbf{I}_\text{min})$, where $\mathbf{I}_\text{max}$ and $\mathbf{I}_\text{min}$ are the maximum and minimum values of $\mathbf{I}$, respectively. Note that, our shadow mask $\mathbf{M}_s$ is a soft map and have the values in the range of $[0, 1]$. See Fig.~\ref{fig:smoothloss_mask} for some examples. 

The noise-robust function $\text{B}$ is defined as: $\text{B}(\mathbf{M}_s) =$ ${\mathbf{B}_{s}}_x + {\mathbf{B}_{s}}_y$ where ${\mathbf{B}_{s}}_x(\text{p}) = \big\lvert\sum_{\text{q}\in \mathbf{R}_\text{p}}g_{\text{p},\text{q}}\partial_x(\mathbf{M}_s(\text{q}))\big\rvert$ and 
${\mathbf{B}_{s}}_y(\text{p}) = \big\lvert\sum_{\text{q}\in \mathbf{R}_\text{p}}g_{\text{p},\text{q}}\partial_y(\mathbf{M}_s(\text{q}))\big\rvert$,  $\partial_x$ and $\partial_y$ are partial derivatives in horizontal and vertical directions respectively, $\text{p}$ defines a pixel, $\mathbf{R}_\text{p}$ is a 3$\times$3 window around $\text{p}$, and $g_{\text{p},\text{q}}$ is a weighing function measuring spatial affinity defined as $g_{\text{p},\text{q}}=\exp\big(\frac{-(\text{p}-\text{q})^2}{2\tau^2}\big)$, where $\tau$ is set to 0.01 by default. See Fig.~\ref{fig:smoothloss}(c) for some examples of our soft boundary detection. 

%-------------------------------------------------------------------------
\vspace{0.2cm}
\subsection{Adversarial, Consistency and Identity Losses}
For shadow removal, we use the generator $G_s$, which is coupled with a discriminator $D_{sf}$. To ensure reconstruction consistency, we use another generator $G_{sf}$ coupled with its own discriminator $D_{s}$. We use adversarial losses to train our DC-ShadowNet:
\begin{align}
	\mathcal{L}_{\rm adv}(G_s, D_{sf})=  
	\mathbb{E}_{\mathbf{I}_{sf}} \big[&\log \big(D_{sf}(\mathbf{I}_{sf})\big)\big]\label{eq:loss_gan1}
	+\\\mathbb{E}_{\mathbf{I}_s} \big[&\log \big(1-D_{sf}(G_s(\mathbf{I}_s))\big)\big],\nonumber
	\\
	\mathcal{L}_{\rm adv}(G_{sf}, D_s)=  
	\mathbb{E}_{\mathbf{I}_{s}} \big[&\log \big(D_s(\mathbf{I}_{s})\big)\big]\label{eq:loss_gan2}
	+ \\ \mathbb{E}_{\mathbf{I}_{sf}}\big[&\log \big(1-D_s(G_{sf}(\mathbf{I}_{sf}, \mathbf{M}_s))\big)\big].\nonumber
\end{align}
During training, the losses expressed in Eqs.~(\ref{eq:loss_gan1}) and (\ref{eq:loss_gan2}) are actually minimized as $\min_{G_s}\max_{D_{sf}}$ $(\mathcal{L}_{\rm adv}(G_s, D_{sf}))$ and $\min_{G_{sf}}\max_{D_s}$ $(\mathcal{L}_{\rm adv}(G_{sf}, D_s))$ respectively. 
Note that, unlike generator $G_s$, the generator $G_{sf}$  takes the mask $\mathbf{M}_s$ (from Eq.~\ref{eq:mask}) as input to help render  more proper shadow images~\cite{Hu19}. Following~\cite{Zhu17,yan2020optical}, we define our reconstruction consistency losses by:
\begin{align}
	\mathcal{L}_{\rm cons}(G_s)&=  \mathbb{E}_{\mathbf{I}_s}\big[||G_{sf}\big(G_s(\mathbf{I}_s), \mathbf{M}_s\big)-\mathbf{I}_{s}||_{1}\big],  
	\label{eq:loss_cycle1}
	\\
	\mathcal{L}_{\rm cons}(G_{sf})& =  \mathbb{E}_{\mathbf{I}_{sf}}\big[||G_s\big(G_{sf}(\mathbf{I}_{sf}, \mathbf{M}_s)\big)-\mathbf{I}_{sf}||_{1}\big].
	\label{eq:loss_cycle2}
\end{align}
While our $G_s$ is designed to remove shadows from shadow input image $\mathbf{I}_s$, we also encourage it to output the same image as input, if the input is a shadow-free image $\mathbf{I}_{sf}$. We achieve this by using the following identity losses~\cite{Zhu17}:
\begin{align}
	\mathcal{L}_{\rm iden}(G_s) &=  
	\mathbb{E}_{\mathbf{I}_{sf}} \big[||(G_s(\mathbf{I}_{sf}))-\mathbf{I}_{sf}||_{1}\big],
	\label{eq:loss_identity1}
	\\
	\mathcal{L}_{\rm iden}(G_{sf}) &= 
	\mathbb{E}_{\mathbf{I}_{s}} \big[||(G_{sf}(\mathbf{I}_s,\mathbf{M}_0)-\mathbf{I}_s)||_{1}\big].
	\label{eq:loss_identity2}
\end{align}
where $\mathbf{M}_0$ represents a mask with all zero values. 

\vspace{0.3cm}
\noindent \textbf{Overall Loss}
We multiply each loss function with its respective weight, and sum them together to obtain our overall loss function. The weights of the losses, \{$\mathcal{L}_{\rm chroma}$,$\mathcal{L}_{\rm feature}$, $\mathcal{L}_{\rm smooth}$, $\mathcal{L}_{\rm domcls}$, $\mathcal{L}_{\rm adv}$, $\mathcal{L}_{\rm cons}$, $\mathcal{L}_{\rm iden}$\}, are represented by \{$\lambda_\text{chroma}$, $\lambda_\text{feat}$, $\lambda_\text{sm}$, $\lambda_\text{dom}$, $\lambda_\text{adv}$, $\lambda_\text{cons}$, $\lambda_\text{iden}$\}.

\begin{table}[t!]
	\small
	\centering
	\thickmuskip=3mu
	\renewcommand{\arraystretch}{1.2}
	\caption {RMSE results on the SRD dataset. All, S and NS represent entire, shadow and non-shadow regions respectively.}
	\label{tb1:srd}
	\begin{tabularx}{\columnwidth}{ c|c|Y|Y|Y }
		\toprule
		Method                      &Training  &All    &S &NS\\
		%\midrule
		\hline
		\bf Our DC-ShadowNet        &Unpaired      &\bf4.66 &7.70 &3.39\\
		%\hline
		Mask-ShadowGAN \cite{Hu19}  &Unpaired      &6.40   &11.46  &4.29\\
		\hline
		DSC~\cite{Hu18}				&Paired        &4.86   &8.81  &\bf3.23\\
		%\hline
		DeShadowNet~\cite{Qu17}		&Paired		   &5.11   &\bf3.57   &8.82\\
		\hline
		Gong~\etal \cite{Gong14}    &-             &12.35  &25.43  &6.91\\
		\hline
		Input Image                 &-             &13.77 &37.40  &3.96\\
		\bottomrule
	\end{tabularx}	
\end{table}
%-------------------------------------------------------------------------
\begin{table}[t!]
	\small
	\centering
	\thickmuskip=3mu
	\renewcommand{\arraystretch}{1.2}
	\caption {RMSE results on the AISTD dataset. All, S and NS represent entire, shadow and non-shadow regions respectively. M shows that ground truth shadow masks are also used in training.}
	\label{tb2:aistd}
	\begin{tabularx}{\columnwidth}{ c|c|Y|Y|Y }
		\toprule
		Method                       &Training &All &S &NS\\
		\hline
		\bf Our DC-ShadowNet 		 &Unpaired &\bf4.6 &\bf10.3 &3.5\\
		%\hline
		Mask-ShadowGAN \cite{Hu19}   &Unpaired &5.3 &12.5 &4.0\\
		\hline
		DeshadowNet~\cite{Qu17}      &Paired   &7.6 &15.9 &6.0\\
		%\hline
		ST-CGAN~\cite{Wang18}        &Paired+M &8.7 &13.4 &7.7\\
		%\hline
		\hline
		Gong \etal \cite{Gong14}     &-        &-  &13.3 &-\\
		%\hline
		Guo \etal \cite{Guo11}       &Paired+M &6.1  &22.0 &3.1\\
		%\hline
		Yang \etal \cite{Yang12}     &-        &16.0 &24.7 &14.4\\
		\hline
		Input Image                  &-        &8.5  &40.2 &\bf2.6\\
		\bottomrule
	\end{tabularx}	
\end{table}
%-------------------------------------------------------------------------

%-------------------------------------------------------------------------
\section{Experiments}
\label{sec:experiments}
\begin{figure}[t]
	\centering
	%\captionsetup[subfigure]{font=small, labelformat=empty}
	\captionsetup[subfloat]{farskip=2pt}
	\setcounter{subfigure}{0}
	\subfloat[Input]
	{\includegraphics[width = 0.162\columnwidth]{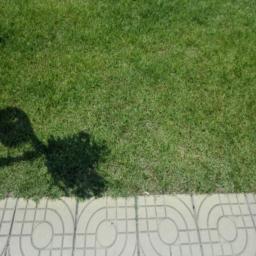}}\hfill
	\subfloat[\textbf{Ours}]
	{\includegraphics[width = 0.162\columnwidth]{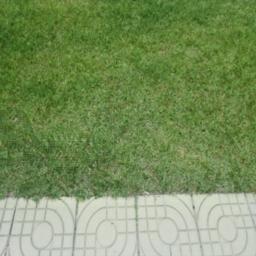}}\hfill
	\subfloat[~\cite{Hu19}]
	{\includegraphics[width = 0.162\columnwidth]{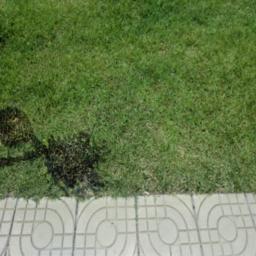}}\hfill
	\subfloat[~\cite{le2020shadow}]
	{\includegraphics[width = 0.162\columnwidth]{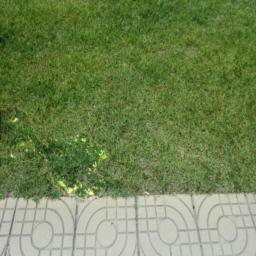}}\hfill
	\subfloat[~\cite{Wang18}]
	{\includegraphics[width = 0.162\columnwidth]{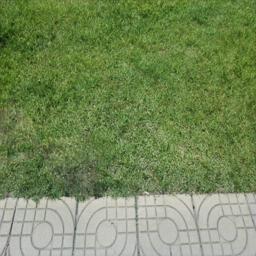}}\hfill	
	\subfloat[~\cite{Gong14}]
	{\includegraphics[width = 0.162\columnwidth]{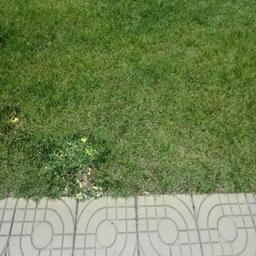}}\hfill\\	
	\vspace{-0.02in}
	\subfloat{\includegraphics[width = 0.162\columnwidth]{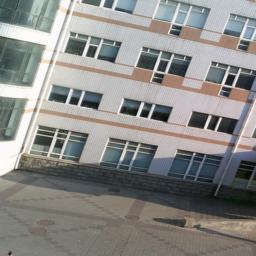}}\hfill
	\subfloat{\includegraphics[width = 0.162\columnwidth]{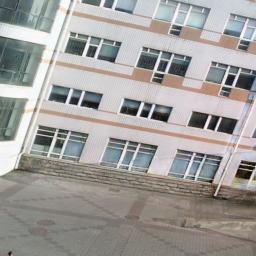}}\hfill
	\subfloat{\includegraphics[width = 0.162\columnwidth]{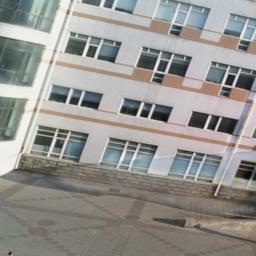}}\hfill
	\subfloat{\includegraphics[width = 0.162\columnwidth]{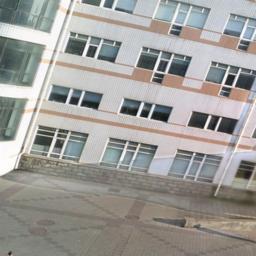}}\hfill	
	\subfloat{\includegraphics[width = 0.162\columnwidth]{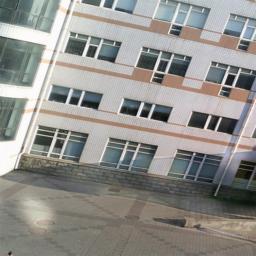}}\hfill	
	\subfloat{\includegraphics[width = 0.162\columnwidth]{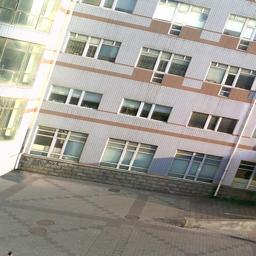}}\hfill\\
	\vspace{-0.02in}	
	\subfloat{\includegraphics[width = 0.162\columnwidth]{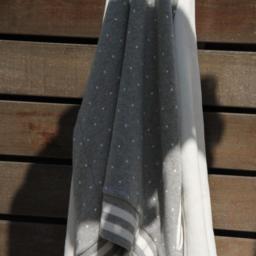}}\hfill
	\subfloat{\includegraphics[width = 0.162\columnwidth]{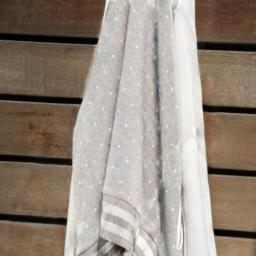}}\hfill
	\subfloat{\includegraphics[width = 0.162\columnwidth]{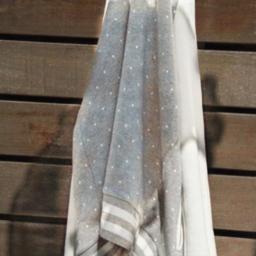}}\hfill
	\subfloat{\includegraphics[width = 0.162\columnwidth]{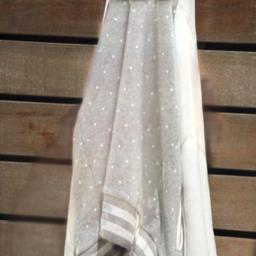}}\hfill
	\subfloat{\includegraphics[width = 0.162\columnwidth]{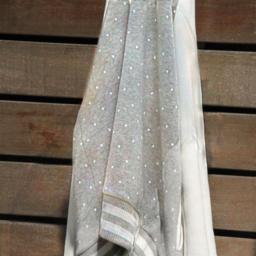}}\hfill
	\subfloat{\includegraphics[width = 0.162\columnwidth]{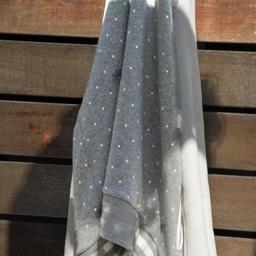}}\hfill\\
	\vspace{-0.02in}
	\setcounter{subfigure}{0}
	\subfloat[Input]
	{\includegraphics[width = 0.162\columnwidth]{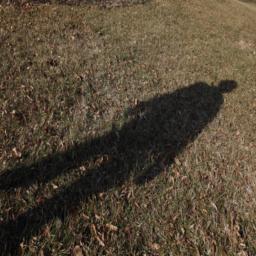}}\hfill
	\subfloat[\textbf{Ours}]
	{\includegraphics[width = 0.162\columnwidth]{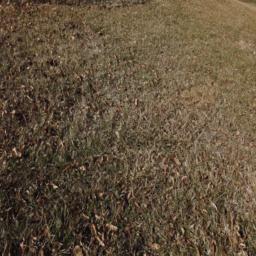}}\hfill
	\subfloat[~\cite{Hu19}]
	{\includegraphics[width = 0.162\columnwidth]{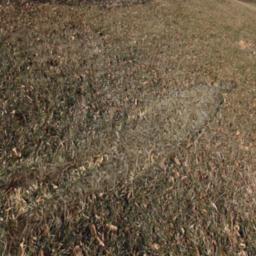}}\hfill
	\subfloat[~\cite{Hu18}]
	{\includegraphics[width = 0.162\columnwidth]{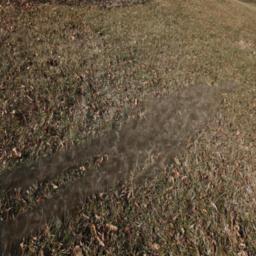}}\hfill	
	\subfloat[~\cite{Qu17}]
	{\includegraphics[width = 0.162\columnwidth]{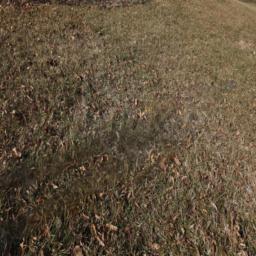}}\hfill	
	\subfloat[~\cite{Gong14}]
	{\includegraphics[width = 0.162\columnwidth]{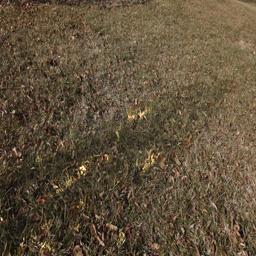}}\hfill\\
	\vspace{-0.01in}
	\caption{\textbf{Comparison results on the ISTD (top row) and SRD (bottom three rows) datasets.} (a) Input image, (b) Our method, unsupervised
	method (c) Mask-ShadowGAN~\cite{Hu19}, weakly-supervised method (d) Param+M+D-Net~\cite{le2020shadow} (top row), supervised methods DSC~\cite{Hu18}, (e) ST-CGAN~\cite{Wang18} (top row), DeshadowNet~\cite{Qu17}, and traditional method (f) Gong~\etal~\cite{Gong14}. Our method trained using unsupervised learning provides the best performance.}
	\vspace{-0.1in}
	\label{fig:SRD}
\end{figure}

%-------------------------------------------------------------------------
To evaluate our method, we use five datasets: SRD~\cite{Qu17}, adjusted ISTD (AISTD)~\cite{Le19}, ISTD~\cite{Wang18}, USR~\cite{Hu19} and LRSS~\cite{Gryka15}, where LRSS is a soft shadow dataset.
To ensure fair comparisons, all the unsupervised baselines, including ours are trained and tested on the same datasets.
For the SRD dataset, for Table~\ref{tb1:srd} and Fig.~\ref{fig:SRD} rows 2-4, we use 2680 shadow images and 2680 shadow-free images for training. We use 408 shadow images that have shadow-free ground truth for testing. 
Similarly, for Table~\ref{tb2:aistd}, we use 1330 training and 540 testing AISTD images; Fig.~\ref{fig:SRD} row 1, we use 1330 training and 540 testing ISTD images. 
For the USR dataset, we use 1956 shadow, 1770 shadow-free images for training, 489 shadow images for testing.
However, for testing, the USR dataset does not provide paired shadow and shadow-free images.

Our DC-ShadowNet is trained in an unsupervised manner (Sec.~\ref{sec:method}). The weights of our losses $\{\lambda_\text{chroma},\lambda_\text{feat},\lambda_\text{sm},\lambda_\text{iden}, \lambda_\text{adv},\lambda_\text{cons},\lambda_\text{dom}\}$ are set empirically to $\{1,1,1,10,1,10,1\}$.
Following the baselines~\cite{Guo12,Hu19}, to evaluate shadow removal performance\footnote{Results of~\cite{Hu19,Hu18,Wang18,Gong14,Le19,Cun20} are taken from their official implementations. Results of~\cite{Gryka15,Guo12} are obtained from their project website: http://visual.cs.ucl.ac.uk/pubs/softshadows/. The quantitative results are taken from the paper~\cite{le2020shadow}.}, we use root mean squared error (RMSE) between the ground truth and the predicted shadow-free image\footnote{As mentioned in~\cite{gitcoderef}, the default RMSE evaluation code used by all methods (including ours) actually computes mean absolute error (MAE).}. 
Hence, lower numbers show better performance.

\begin{table}
	\centering
	\caption{
	RMSE (lower is better) and PSNR (higher is better) results on the LRSS dataset (soft shadow dataset). M and S respectively show that ground truth shadow masks and synthetic paired data are used in training. P and UP denote paired and unpaired training, respectively.
	}
	\vspace{0.5mm}
	\resizebox{1.0\columnwidth}{!}{
		\begin{tabular}{c|c|ccc|cc|cc}
			\toprule
			Method  &Input  &\cite{Guo12} &\cite{Guo12} \footnotesize(auto)  &\cite{Gryka15} &\cite{Cun20}  &\cite{Le19} &\cite{Hu19} &\bf Ours \\
			%\midrule
			\hline
			RMSE    &12.26 &6.02   &5.87    &4.38       &7.92 &7.48    		&7.13       &\bf 3.48\\
			%\hline
			PSNR    &18.05 &27.88  & 28.02  &29.25      &25.57&23.93   		&25.12      &\bf 31.01\\
			%\midrule
			\hline
			Training&-     &P+M    &P       &P+M+S      &P+M+S      &P+M  &UP &UP\\				
			\bottomrule
		\end{tabular}
	}
	\vspace{-0.1cm}
	\label{tb3:lrss}
\end{table}

\vspace{0.2cm}
\noindent{\bf Results on Hard Shadows}
We conduct quantitative evaluations on the SRD and AISTD datasets, and the corresponding results are shown in Table~\ref{tb1:srd} and Table~\ref{tb2:aistd}, respectively.
For comparisons, we use the state-of-the-art unsupervised shadow removal method Mask-ShadowGAN~\cite{Hu19}, weakly-supervised method Param+M+D-Net~\cite{le2020shadow}, supervised methods DSC~\cite{Hu18}, DeshadowNet~\cite{Qu17}, ST-CGAN~\cite{Wang18},
and traditional methods Gong \etal.~\cite{Gong14}, Guo \etal.~\cite{Guo12}, and Yang \etal.~\cite{Yang12}.
From Tables~\ref{tb1:srd} and \ref{tb2:aistd}, our DC-ShadowNet trained in an unsupervised manner achieves the best performance compared to the baseline methods. 
Compared to the state-of-the-art unsupervised method Mask-ShadowGAN~\cite{Hu19}, our results for the shadow regions are better by $\sim$33\% and $\sim$18\% on the SRD and AISTD datasets, respectively.
%------------------------------------------------------------------------

The qualitative results for the SRD (rows 2-4) and ISTD (top row) datasets are shown in Fig.~\ref{fig:SRD}, which include challenging conditions and diverse objects. For example, the shadow image contains shadows casted on semantic objects (i.e., building, wall).
In Fig.~\ref{fig:SRD}, the method~\cite{Hu19} alters the colors of the non-shadow regions and cannot properly handle shadow boundaries. For the method~\cite{Gong14}, the recovery of the shadow-free images is unsatisfactory. 
In comparison, our DC-ShadowNet performs better, showing the effectiveness of our domain classification network and our novel unsupervised losses.

%------------------------------------------------------------------------
\begin{table}[t!]
	\small
	\centering
	\thickmuskip=3mu
	\renewcommand{\arraystretch}{1.2}
	\caption {Ablation experiments of our method using the SRD dataset. All, S and NS represent entire, shadow and non-shadow regions, respectively. The numbers represent RMSE.}	
	\label{tb:ablation}
	\begin{tabularx}{\columnwidth}{ c|Y|Y|Y }
		\toprule
		Method							&All &S &NS\\
		%\midrule
		\hline
		\bf Our DC-ShadowNet 			&\bf4.66 &\bf7.70  &\bf3.39\\	
		\hline
		w/o $\mathcal{L}_{\rm smooth}$  &4.72 &7.80 &3.43\\
		%\hline
		w/o $\mathcal{L}_{\rm feature}$ &4.83 &8.04  &3.50\\
		%\hline
		w/o $\mathcal{L}_{\rm chroma}$  &5.05 &8.50  &3.61\\
		%\hline	
		w/o $\Phi^g_s$       	        &5.20 &8.94  &3.65\\
		%\hline	
		w/o $\Phi^d_{sf}$       	    &5.49 &9.42  &3.87\\
		%\hline
		w/o $\Phi^g_s$ and $\Phi^d_{sf}$   &8.12 &16.10  &4.80\\
		\hline
		Input Image                 	&13.77 &37.40 &3.96\\
		\bottomrule	
	\end{tabularx}
\end{table}

%-------------------------------------------------------------------------
\begin{figure*}[t]
	\centering
	\vspace{-0.1in}
	%\captionsetup[subfigure]{font=small, labelformat=empty}
	\captionsetup[subfloat]{farskip=2pt}
	\subfloat{\includegraphics[width = 0.254\columnwidth]{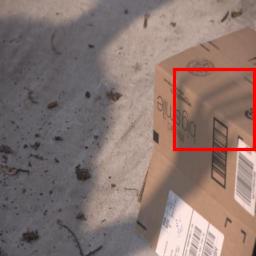}}\hfill
	\subfloat{\includegraphics[width = 0.254\columnwidth]{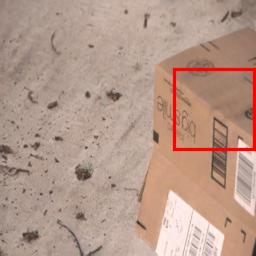}}\hfill
	\subfloat{\includegraphics[width = 0.254\columnwidth]{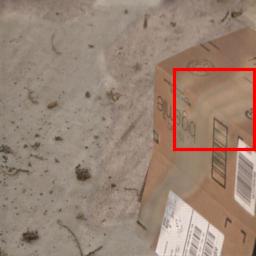}}\hfill
	\subfloat{\includegraphics[width = 0.254\columnwidth]{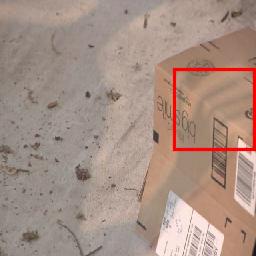}}\hfill
	\subfloat{\includegraphics[width = 0.254\columnwidth]{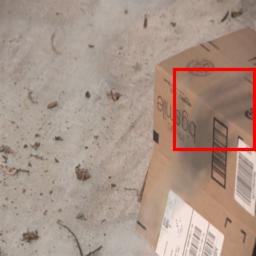}}\hfill
	\subfloat{\includegraphics[width = 0.254\columnwidth]{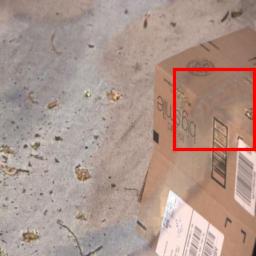}}\hfill
	\subfloat{\includegraphics[width = 0.254\columnwidth]{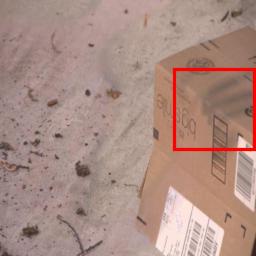}}\hfill
	\subfloat{\includegraphics[width = 0.254\columnwidth]{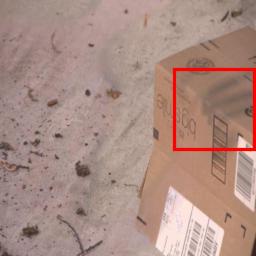}}\hfill\\
	\vspace{-0.02in}
	\subfloat{\includegraphics[width = 0.254\columnwidth]{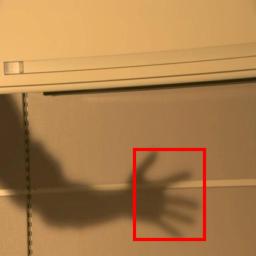}}\hfill
	\subfloat{\includegraphics[width = 0.254\columnwidth]{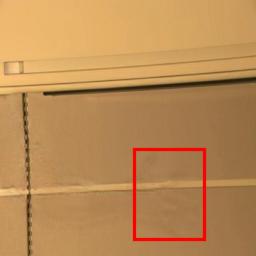}}\hfill
	\subfloat{\includegraphics[width = 0.254\columnwidth]{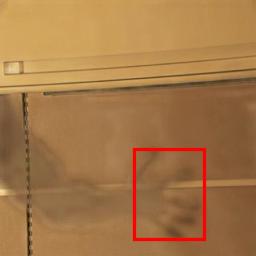}}\hfill
	\subfloat{\includegraphics[width = 0.254\columnwidth]{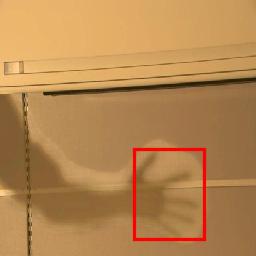}}\hfill
	\subfloat{\includegraphics[width = 0.254\columnwidth]{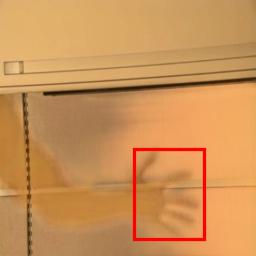}}\hfill
	\subfloat{\includegraphics[width = 0.254\columnwidth]{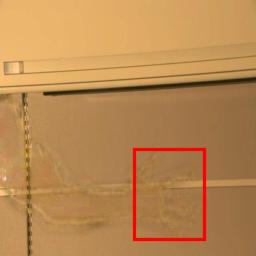}}\hfill
	\subfloat{\includegraphics[width = 0.254\columnwidth]{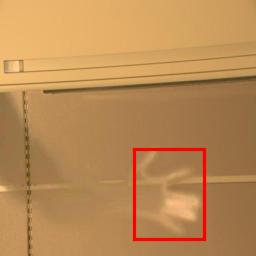}}\hfill
	\subfloat{\includegraphics[width = 0.254\columnwidth]{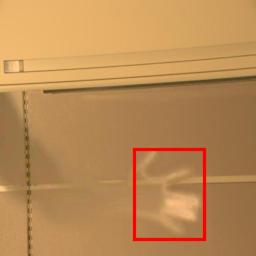}}\hfill\\
	\vspace{-0.02in}
	\subfloat{\includegraphics[width = 0.254\columnwidth]{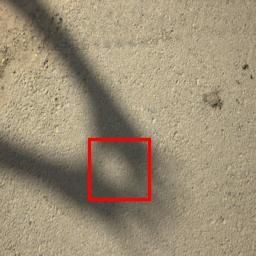}}\hfill
	\subfloat{\includegraphics[width = 0.254\columnwidth]{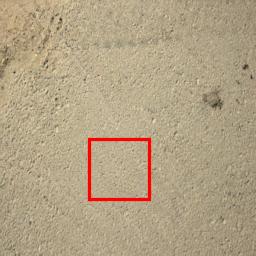}}\hfill
	\subfloat{\includegraphics[width = 0.254\columnwidth]{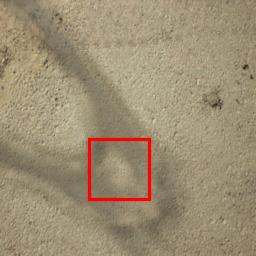}}\hfill
	\subfloat{\includegraphics[width = 0.254\columnwidth]{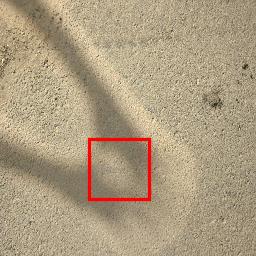}}\hfill
	\subfloat{\includegraphics[width = 0.254\columnwidth]{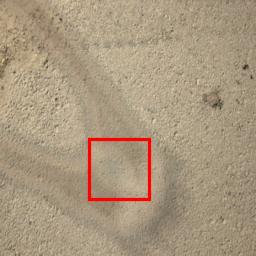}}\hfill
	\subfloat{\includegraphics[width = 0.254\columnwidth]{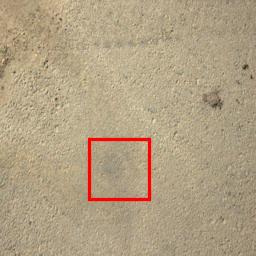}}\hfill
	\subfloat{\includegraphics[width = 0.254\columnwidth]{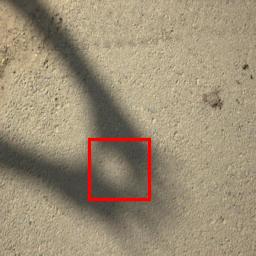}}\hfill
	\subfloat{\includegraphics[width = 0.254\columnwidth]{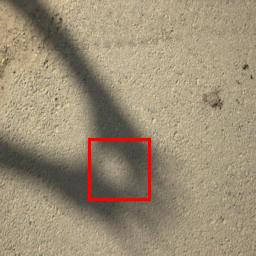}}\hfill\\
	\vspace{-0.02in}
	\subfloat{\includegraphics[width = 0.254\columnwidth]{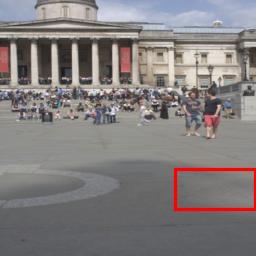}}\hfill
	\subfloat{\includegraphics[width = 0.254\columnwidth]{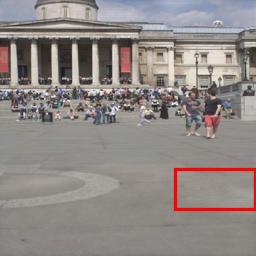}}\hfill
	\subfloat{\includegraphics[width = 0.254\columnwidth]{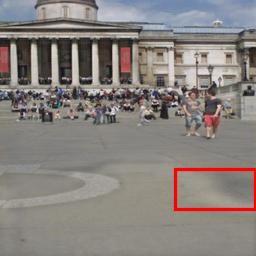}}\hfill
	\subfloat{\includegraphics[width = 0.254\columnwidth]{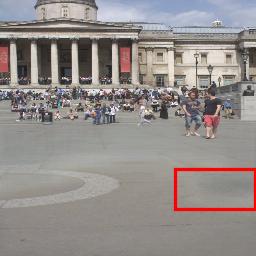}}\hfill
	\subfloat{\includegraphics[width = 0.254\columnwidth]{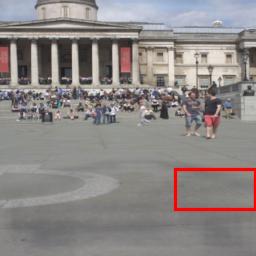}}\hfill
	\subfloat{\includegraphics[width = 0.254\columnwidth]{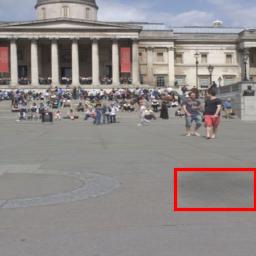}}\hfill
	\subfloat{\includegraphics[width = 0.254\columnwidth]{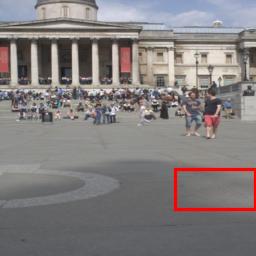}}\hfill
	\subfloat{\includegraphics[width = 0.254\columnwidth]{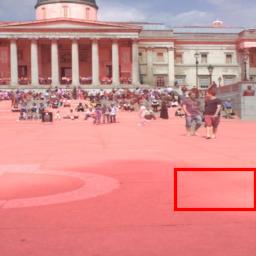}}\hfill\\
	\setcounter{subfigure}{0}
	\subfloat[Input]	
	{\includegraphics[width = 0.254\columnwidth]{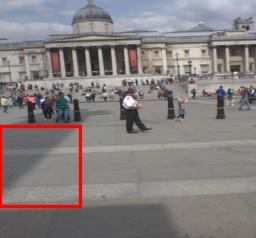}}\hfill
	\subfloat[\textbf{Ours}]
	{\includegraphics[width = 0.254\columnwidth]{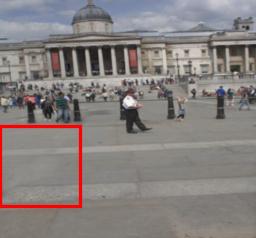}}\hfill
	\subfloat[\cite{Hu19}]	
	{\includegraphics[width = 0.254\columnwidth]{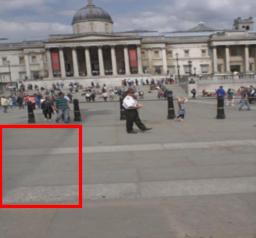}}\hfill
	\subfloat[\cite{Le19}]	
	{\includegraphics[width = 0.254\columnwidth]{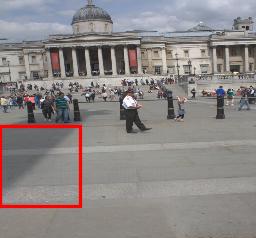}}\hfill
	\subfloat[\cite{Cun20}]
	{\includegraphics[width = 0.254\columnwidth]{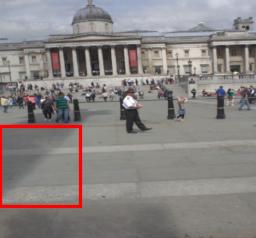}}\hfill
	\subfloat[Gryka~\cite{Gryka15}]
	{\includegraphics[width = 0.254\columnwidth]{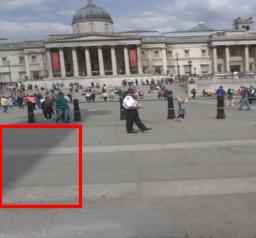}}\hfill
	\subfloat[Guo~\cite{Guo12}]
	{\includegraphics[width = 0.254\columnwidth]{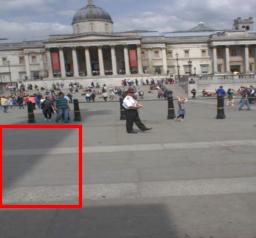}}\hfill
	\subfloat[\cite{Guo12} (auto)]
	{\includegraphics[width = 0.254\columnwidth]{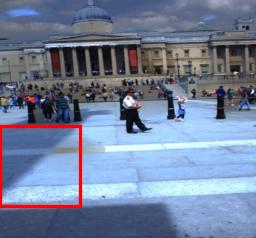}}\hfill\\
	\caption{\textbf{Comparison results on the soft shadow LRSS dataset} (a) Input image, (b) Our result, (c) Unsupervised method Mask-ShadowGAN~\cite{Hu19}, Supervised methods (d) SP+M-Net~\cite{Le19} and (e) DHAN~\cite{Cun20}. (f)$\sim$(h) are the results of the traditional methods (auto means automatic detection). Our method, trained using unsupervised learning, generates better shadow-free results.}
	\label{fig:soft}
\end{figure*}

\vspace{0.2cm}
\noindent{\bf Results on Soft Shadows}
The LRSS dataset has 134 shadow images, mainly contains soft-shadow images. 
We pre-trained our DC-ShadowNet on the SRD training set, then we use 100 LRSS images for training it  in an unsupervised manner. The remaining 34 LRSS images with their corresponding shadow-free images are used for testing.
The quantitative results are shown in Table~\ref{tb3:lrss}.
We compare our DC-ShadowNet with the following methods: unsupervised method Mask-ShadowGAN~\cite{Hu19}, supervised methods SP+M-Net~\cite{Le19} and DHAN~\cite{Cun20}, automatic method Guo~\cite{Guo12}, and interactive method~\cite{Gryka15} which requires user-annotations of shadow regions. 
As shown in Table~\ref{tb3:lrss}, our method achieves the lowest RMSE and highest PSNR.
 
The qualitative results covering a diverse set of images such as indoor/outdoor scenes, shadow regions, etc., are shown in Fig.~\ref{fig:soft}.
While the state-of-the-art methods can remove shadows to some extent, the results are still improper.
Mask-ShadowGAN~\cite{Hu19} fails to handle soft-shadows since it uses binary masks to represent shadow regions. Moreover, it mainly relies on adversarial training that cannot guarantee proper shadow removal. 
Supervised methods like DHAN~\cite{Cun20} and SP+M-Net~\cite{Le19} have artifacts in the shadow regions as they suffer from the domain gap problem.  
Guo~\cite{Guo12} fails due to the difficulty in automatically identifying soft shadow regions.
Compared to all the baseline methods, our results are more proper, and the image surfaces are better-restored. 
%------------------------------------------------------------------------

\begin{figure}[t]
	\centering
	\captionsetup[subfloat]{farskip=2pt}
	\subfloat[Input]
	{\includegraphics[width = 0.245\columnwidth]{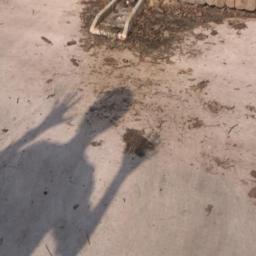}}\hfill
	\subfloat[Ours (w/o)]
	{\includegraphics[width = 0.245\columnwidth]{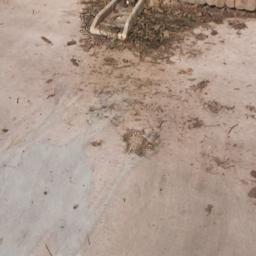}}\hfill
	\subfloat[Ours (w)]
	{\includegraphics[width = 0.245\columnwidth]{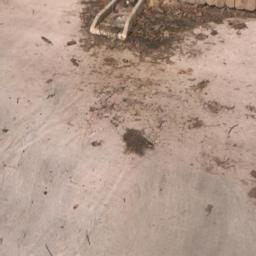}}\hfill
	\subfloat[~\cite{Hu19}]
	{\includegraphics[width = 0.245\columnwidth]{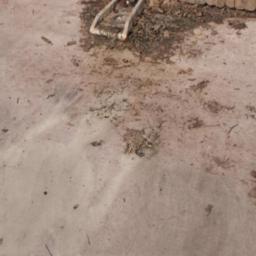}}\hfill\\
	\vspace{-0.01in}
	\caption{ (a) Input image, (b) and (c) show our results without and with test-time-training, (d) Result of Mask-ShadowGAN~\cite{Hu19}.}
	\label{fig:testtime}
\end{figure}

\vspace{0.2cm}
\noindent{\bf Test-Time Training}
We show that our method being unsupervised can be used for test-time training to further improve the results on the test images. For this, we use the 34 shadow images from the test set used in the soft shadow evaluation above, and employ our unsupervised losses to train our method.
To evaluate shadow removal performance, we use the corresponding shadow-free images; and the performance in terms of RMSE and PSNR improves from 3.48 and 31.01 to 3.36 and 31.31, respectively. 
See Fig.~\ref{fig:testtime} for a qualitative example showing the effectiveness of test-time training. 
%------------------------------------------------------------------------
\section{Ablation Study}
\label{sec:ablation}
We conduct ablation studies to analyze the effectiveness of different components of our method such as the shadow-invariant chromaticity loss $\mathcal{L}_{\rm chroma}$, shadow-robust feature loss $\mathcal{L}_{\rm feature}$, boundary-smoothness loss $\mathcal{L}_{\rm smooth}$, and the domain classifier $\Phi^g_s$ and $\Phi^d_{sf}$. We use the SRD dataset for our experiments and the corresponding quantitative results are shown in Table~\ref{tb:ablation}. Each component of our method is important and contributes to the better performance.

\section{Conclusion}
\label{sec:conclusion}
We have proposed DC-ShadowNet, an unsupervised learning-based shadow removal method guided by domain classification network, shadow-free chromaticity, shadow-robust feature and boundary smoothness losses. 
Our method can robustly handle both hard and soft shadow images. 
We integrate a domain classifier with our generator and its corresponding discriminator, enabling our method to focus on shadow regions.
To train DC-ShadowNet, we use novel unsupervised losses that enable it to directly learn from unlabeled (no ground truth) real shadow images.   
We also showed that we could employ test-time refinement that can further improve our performance. 
Experimental results have confirmed that our method is effective and outperforms the state-of-the-art shadow removal methods.

{\small
	\bibliographystyle{ieee_fullname}
	\bibliography{egbib}
}

\end{document}